\pdfoutput=1

\documentclass[11pt]{article}

\usepackage[preprint]{acl}
\usepackage{xspace}
\usepackage{times}
\usepackage{latexsym}
\usepackage{amsmath}
\usepackage{booktabs}
\usepackage{multirow} 
\usepackage{amssymb}
\usepackage{amsmath}
\usepackage{graphicx}
\usepackage{booktabs}
\usepackage{makecell}
\usepackage{subfigure} 
\usepackage{array} 
\usepackage{bbding}
\usepackage{bm}
\usepackage{amssymb}
\usepackage{pifont}
\usepackage{url}
\usepackage{amsthm}

\usepackage{algorithm}
\usepackage{algorithmic}

\newcommand\webqsp{{\sc WebQuestionsSP}\xspace}
\newcommand\cwq{{\sc ComplexWebQuestions}\xspace}
\newcommand\wtq{{\sc WikiTableQuestions}\xspace}

\usepackage[T1]{fontenc}

\usepackage[utf8]{inputenc}

\usepackage{microtype}

\usepackage{inconsolata}

\usepackage{graphicx}

\usepackage{colortbl}

\definecolor{blue3}{RGB}{255,255,255}
\definecolor{blue2}{RGB}{242,249,255}
\definecolor{blue1}{RGB}{225,242,250}
\definecolor{blue0}{RGB}{204,228,249}

\title{SKA-Bench: A Fine-Grained Benchmark for Evaluating Structured Knowledge Understanding of LLMs}

\author{Zhiqiang Liu$^{1,3}$, Enpei Niu$^{1,3}$, Yin Hua$^{1,3}$, Mengshu Sun$^4$, Lei Liang$^4$, \\ {\bf Huajun Chen$^{2,3}$, Wen Zhang$^{1,3}$\thanks{Corresponding Author.}
}\\
$^1$School of Software Technology, Zhejiang University \\
$^2$College of Computer Science and Technology, Zhejiang University\\
$^3$ZJU-Ant Group Joint Lab of Knowledge Graph \\
$^4$Ant Group \\
\texttt{\{zhiqiangliu,zhang.wen\}@zju.edu.cn} \\}

\begin{document}
\maketitle
\begin{abstract}
Although large language models (LLMs) have made significant progress in understanding Structured Knowledge (SK) like KG and Table, existing evaluations for SK understanding are non-rigorous (i.e., lacking evaluations of specific capabilities) and focus on a single type of SK. Therefore, we aim to propose a more comprehensive and rigorous structured knowledge understanding benchmark to diagnose the shortcomings of LLMs. In this paper, we introduce \textbf{\textit{SKA-Bench}}, a \textbf{\underline{S}}tructured \textbf{\underline{K}}nowledge \textbf{\underline{A}}ugmented QA \textbf{\underline{Bench}}mark that encompasses four widely used structured knowledge forms: KG, Table, KG+Text, and Table+Text. We utilize a three-stage pipeline to construct \textit{SKA-Bench} instances, which includes a question, an answer, positive knowledge units, and noisy knowledge units. To evaluate the SK understanding capabilities of LLMs in a fine-grained manner, we expand the instances into four fundamental ability testbeds: \textit{Noise Robustness}, \textit{Order Insensitivity}, \textit{Information Integration}, and \textit{Negative Rejection}. Empirical evaluations on 8 representative LLMs, including the advanced DeepSeek-R1, indicate that existing LLMs still face significant challenges in understanding structured knowledge, and their performance is influenced by factors such as the amount of noise, the order of knowledge units, and hallucination phenomenon. Our dataset and code are available at \url{https://github.com/zjukg/SKA-Bench}.
\end{abstract}

\section{Introduction}
With the rapid development of large language mo-

\noindent dels (LLMs)~\cite{gpt4,llama3}, Structured Knowledge (SK), such as knowledge graphs (KG)~\cite{freebase} and tables, still remain essential due to their systematic and rigorous organizational formats. On the one hand, structured knowledge is usually present in various real-world scenarios (e.g., financial reports with numerous tables~\cite{finqa} and product knowledge graphs~\cite{stark}), thus serving as a significant knowledge base for existing LLM systems~\cite{kag,pikerag}. On the other hand, due to their well-organized structure and intensive knowledge characteristics, structured knowledge is also widely utilized to improve the inference-time performances of LLMs~\cite{chainofknowledge,structrag,AAAI24}. Consequently, evaluating the ability of LLMs to understand structured knowledge is a crucial research topic. 

\begin{figure}
\centering
\includegraphics[scale=0.75]{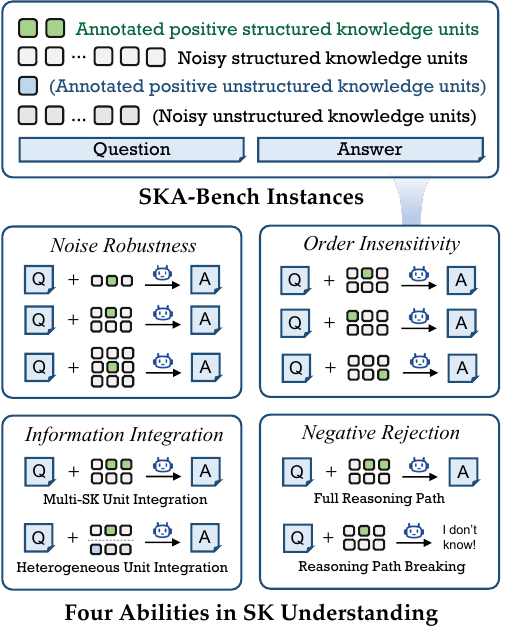}
\vspace{-1mm}
\caption{The components of a \textit{SKA-Bench} instance and how to further construct the four ability testbeds for evaluating structured knowledge understanding.} \label{intro}
\vspace{-1mm}
\end{figure}

Unlike common unstructured text understanding tasks~\cite{evaluate-survey,chen2024large}, LLMs still face significant challenges~\cite{tableqa-survey,ontotune} in understanding structured knowledge. This is because LLMs need to capture long-distance contextual dependencies as well as complex relationships and hierarchical structures from the given structured knowledge. However, existing benchmarks~\cite{wtq,tablebench,cwq,stark} for evaluating structured knowledge understanding suffer from limitations, including the lack of detailed reasoning path annotations or sufficiently long structured knowledge bases, making it difficult to thoroughly diagnose the shortcomings of LLMs in structured knowledge understanding. Moreover, these datasets primarily focus on single data types, including tables~\cite{wtq,tablebench}, knowledge graphs~\cite{cwq}, or hybrid~\cite{hybridqa,stark} formats, which restrict their coverage and fail to fully reflect the comprehensive understanding abilities of the models. \textit{Therefore, there is an urgent need for a diverse and fine-grained dataset to comprehensively evaluate LLMs and identify potential bottlenecks in their structured knowledge understanding capabilities.}

To this end, we construct a fine-grained \textbf{\underline{S}}tru-ctured \textbf{\underline{K}}nowledge \textbf{\underline{A}}ugmented QA \textbf{\underline{Bench}}mark, \textbf{\textit{SKA-Bench}}, which consists of 921 SKA-QA instances and covers four widely used types of structured data. To ensure the quality and complexity of the instances, we propose a novel three-stage construct pipeline for precise positive knowledge unit annotation and the synthesis of long structured knowledge. As illustrated in Fig.~\ref{intro}, \textit{SKA-Bench} instances are composed of a question, an answer, positive knowledge units, and noisy knowledge units, which endow \textit{SKA-Bench} with strong scalability. Ultimately, based on the different compositions of SK units as the given structured knowledge bases, we expand these instances into four distinct testbeds, each targeting a fundamental capability required for understanding SK: \textit{Noise Robustness}, \textit{Order Insensitivity}, \textit{Information Integration}, and \textit{Negative Rejection} for comprehensively diagnosing the shortcomings of LLMs in SK understanding. 

We conduct empirical evaluations on 8 representative LLMs. Even advanced LLMs like DeepSeek-R1 continue to face challenges in SK understanding, with their performance significantly influenced by the amount of noise and the order of knowledge units. Moreover, its negative rejection ability is even weaker than that of certain LLMs with 7B parameters. We hope that \textit{SKA-Bench} can serve as a comprehensive and rigorous benchmark to accelerate the progress of LLMs in understanding and reasoning over structured knowledge.  

\section{Related Work}
\paragraph{Evaluation for Structured Knowledge Understanding.} Current structured knowledge understanding evaluations often focus on knowledge graphs~\cite{webqsp,cwq,graphqa} and tables~\cite{wtq,wikisql,tablebench}. Earlier Table QA datasets, such as WTQ~\cite{wtq}, WikiSQL~\cite{wikisql}, and TabFact~\cite{tabfact} require to retrieve several specific table cells with less than 3 hops, posing limited challenges for LLMs. Recently, \citet{tablebench} propose a more complex Table QA benchmark TableBench for LLM evaluation. However, we believe that the existing evaluations aren't comprehensive enough. On the one hand, the tables in these Table QA datasets are relatively short (average <16.7 rows), making it difficult to evaluate the ability of LLMs to handle long structured knowledge. On the other hand, these datasets lack detailed reasoning path annotations, limiting their utility in fine-grained evaluation of LLMs’ understanding capabilities. For existing KGQA datasets, such as WebQSP~\cite{webqsp}, CWQ~\cite{cwq}, and GraphQA~\cite{graphqa}, they are constructed upon large-scale KGs, thus providing a foundation for creating long and complex KG understanding datasets. But they also lack precise positive triple annotations for systematic evaluation and analysis.

\begin{figure*}
\centering
\includegraphics[scale=0.57]{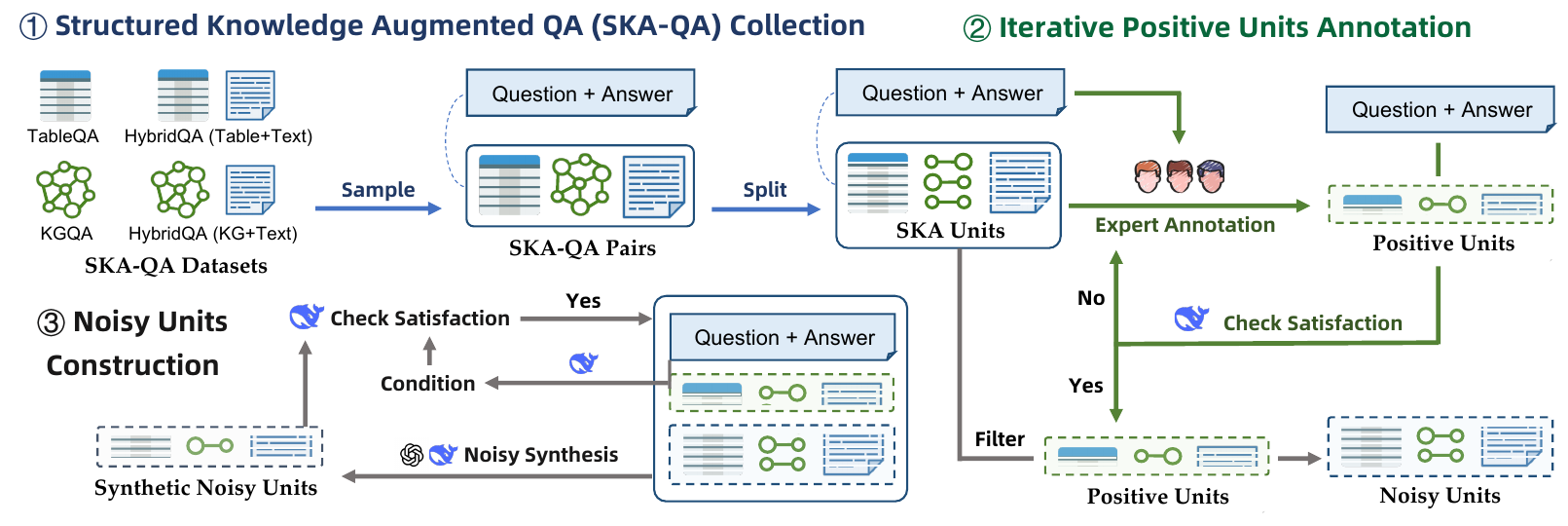}
\vspace{-3mm}
\caption{The construct pipeline to generate \textit{SKA-Bench} instance, which consists of structured knowledge augmented question \& answer (SKA-QA), positive knowledge units and noisy structured units.} \label{overview}
\vspace{-1mm}
\end{figure*}

\paragraph{Evaluation for Semi-structured Knowledge Understanding.}
To more effectively evaluate the understanding of heterogeneous data, the research community has begun to focus on semi-structured knowledge~\cite{hybridqa,tatqa,stark} (i.e., structured data integrated with unstructured textual documents). The semi-structured dataset HybridQA~\cite{hybridqa}, which combines table and textual data, was first proposed. Subsequently, TAT-QA~\cite{tatqa} and FinQA~\cite{finqa} extend the evaluation of understanding and reasoning to more realistic scenarios based on this data format. In addition, STaRK~\cite{stark} dataset based on KG and textual knowledge bases introduces a new retrieval and reasoning challenge for LLMs. However, these hybridQA datasets are also limited by relatively short length of tables or lack of precise annotations, making them challenging for systematic evaluation.

Based on the above considerations, we believe that offering a diverse, fine-grained, and complex benchmark is valuable for thoroughly evaluating LLMs' structured knowledge understanding ability.

\section{SKA-Bench}
\subsection{Problem Definition}
To comprehensively evaluate the ability of LLMs in structured knowledge understanding, \textit{SKA-Bench} incorporates four common types of (semi-)structured data: Knowledge Graph (KG) $\mathcal{G}$, Table $\mathcal{T}$, Knowledge Graph with Textual Documents $\mathcal{G}\cup\mathcal{D}$, and Table with Textual Documents $\mathcal{T}\cup\mathcal{D}$. Following the most existing LLM evaluations~\cite{evaluate-survey1,evaluate-survey}, \textit{SKA-Bench} also adopts a question-answering (QA) format. For a given question $\mathcal{Q}$ and its corresponding structured knowledge $\mathcal{SK}\in$$\{\mathcal{G},\mathcal{T},\mathcal{G}\cup\mathcal{D},\mathcal{T}\cup\mathcal{D}\}$, the LLM $f_{\theta}$ aims to generate the correct answer $\mathcal{A}$, such that $\mathcal{A}=f_{\theta}(\mathcal{Q},\mathcal{SK})$. We hypothesis that LLMs must accurately understand structured knowledge (SK) as a prerequisite for generating correct answers. Therefore, this task format can thoroughly evaluate the SK understanding capabilities of LLMs.  

\subsection{SKA-Bench Construction}
In this section, we detail the construction process of \textbf{\underline{S}}tructured \textbf{\underline{K}}nowledge \textbf{\underline{A}}ugmented \textbf{\underline{Bench}}mark (\textbf{\textit{SKA-Bench}}), which includes three stages: SKA-QA pairs collection, iterative positive units annotation and noisy units synthesis, shown in Fig~\ref{overview}.

\subsubsection{SKA-QA Pairs Collections}
\vspace{0.5mm}
\noindent\textbf{Knowledge Graph.}
We randomly select 900 samples from the test set of KGQA datasets: \webqsp (\textit{WebQSP})~\cite{webqsp} and \cwq (\textit{CWQ})~\cite{cwq} as the initial SKA-QA pairs of \texttt{KG} subset. These two datasets cover 7 common KG relational patterns~\cite{grailqa++} and are both based on widely used Freebase KG~\cite{freebase}. For each QA sample, we extract up to 4-hop subgraph of the topic entities~\cite{unikgqa} in Freebase as the structured knowledge base.

\vspace{0.5mm}
\noindent\textbf{Table.}
We randomly select 700 samples from the widely used Table QA dataset \textit{WTQ}~\cite{wtq} and \textit{TableBench}~\cite{tablebench} with multi-domain, multi-hop question as the initial SKA-QA pairs of \texttt{Table} subset. And our selected tables contain at least 6 columns and 8 rows to facilitate the subsequent synthesis of noisy data. 

\vspace{0.5mm}
\noindent\textbf{KG with Textual Documents.}
We choose the \textit{STaRK}~\cite{stark} dataset, which is constructed based on both textual and relational knowledge bases. Specifically, we randomly select 300 QA samples from both \textit{STaRK-Prime} and \textit{STaRK-Amazon}. For each QA sample, we extract the 2-hop subgraph of the answer entity and the textual descriptions of neighboring nodes within subgraph as the corresponding structured knowledge base. Additionally, we remove SKA-QA pairs where the number of triples in subgraph is less than 200.

\begin{figure*}
\centering
\includegraphics[scale=0.56]{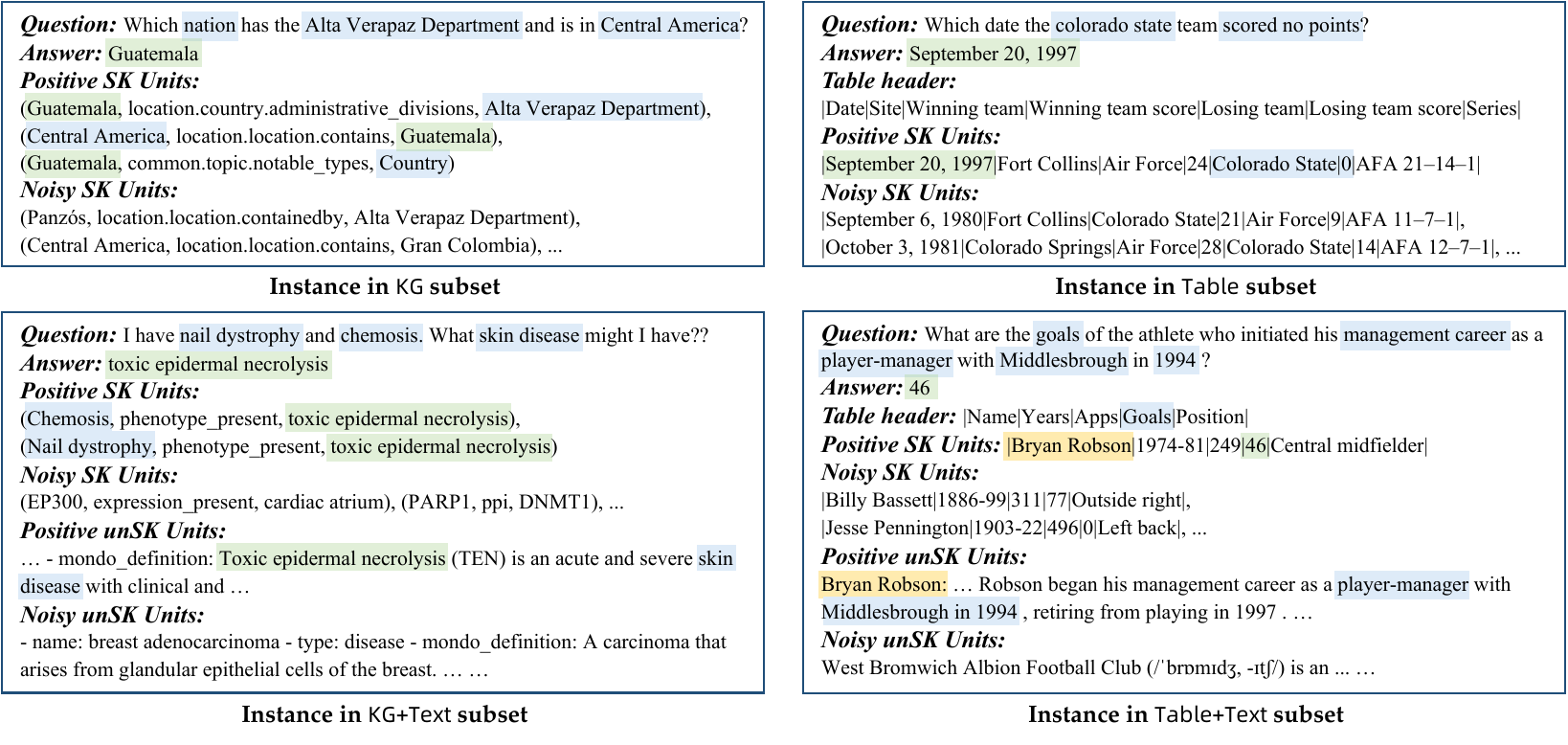}
\vspace{-3mm}
\caption{Four Instances from different subsets of \textbf{\textit{SKA-Bench}}: LLMs need to understand structured knowledge, then select relevant knowledge units to get the answer.} \label{instance}
\vspace{-2mm}
\end{figure*}

\vspace{0.5mm}
\noindent\textbf{Table with Textual Documents.}
For this hybrid data, we also require that QA tasks simultaneously utilize multiple data types. Therefore, we select 200 samples from \textit{HybridQA}~\cite{hybridqa} dataset as a subset. This dataset necessitates reasoning based on heterogeneous knowledge sources and has been widely used in the research community~\cite{qa-survey,tableqa-survey}.

After obtaining the above four types of SKA-QA pairs, we perform a fine-grained split for structured knowledge. Specifically, we regard the triples $\mathcal{F}$ in the KG $\mathcal{G}$ and the rows $\mathcal{R}$ in the tables $\mathcal{T}$ into individual ``\textit{structured knowledge units}'', represented as $\mathcal{G}=\{\mathcal{F}_i\}_{i=1}^n$ and $\mathcal{T}=\mathcal{H}\cup\{\mathcal{R}_j\}_{j=1}^n$. For the table header $\mathcal{H}$, they are separated out independently to preserve the semantic integrity of the table. As for the textual data, we retain the original paragraph-level split in the initial SKA-QA pairs.

\subsubsection{Iterative Positive Units Annotation}
We invite three human experts with computer science backgrounds to perform positive units annotation. Specifically, we require the human experts to accurately identify the positive units required to derive the answer to the given question. Furthermore, the annotation process need to adhere to the following requirements: \textbf{(1)} if the answer is wrong, delete the sample directly; \textbf{(2)} if the question involves multiple answers, all positive units require to obtain the answers should be annotated; \textbf{(3)} for the \texttt{Table} subset and \texttt{Table+Text} subset, if the question needs to perform numerical analysis on the entire table, the corresponding SKA-QA pairs should either be removed or the question should be modified; \textbf{(4)} if the tables in the \texttt{Table} subset and \texttt{Table+Text} subset are order-dependent (i.e., modifying the row order would result in semantic errors in the table), this sample should be removed; \textbf{(5)} for the \texttt{KG+Text} subset and \texttt{Table+Text} subset, if question only utilizes one type of knowledge source, the question should be modified or removed.

After each round of annotation, we query the LLM (utilizing DeepSeek-v3~\cite{deepseek-v3}) to determine whether annotated positive units can derive the answer to the given question. If the response is ``\textit{No}'', re-annotation is performed. The iterative annotation process continues until more than 95\% of the samples receive a ``\textit{Yes}'' response, at which point the iteration is terminated.

\begin{table*}
\centering
\renewcommand{\arraystretch}{1.0}
\resizebox{\textwidth}{!}{
\begin{tabular}{lccccccc}
\toprule
\textbf{Subset}  &\textbf{\#avg $\mathcal{Q}$ token}   &\textbf{\#avg $\mathcal{A}$ num}   &\textbf{\;\#num P (SK/unSK)}    &\textbf{\,\#avg P token\,}    &\textbf{\#num N}  &\textbf{\;\#data\;} &\textbf{Expert Time} \\ \midrule
\textit{\textbf{SKA-Bench}}-\texttt{KG} &15.75   &1.96   &4.25  &16.77  &4541.39  &233 & 5.9 min   \\
\textit{\textbf{SKA-Bench}}-\texttt{Table}  &23.31  &1.10  &3.40  &30.88  &1521.83    &295  &3.6 min  \\
\textit{\textbf{SKA-Bench}}-\texttt{KG+Text}   &30.76    &1.86     &2.53/1.92    &22.31/1053.55     &417.29/79.84    &195 &6.8 min \\ 
\textit{\textbf{SKA-Bench}}-\texttt{Table+Text}  &22.41    &1.01     &1.17/1.28    &28.37/203.78     &1144.55/661.90    &198 &5.8 min   \\ \bottomrule
\end{tabular}}
\vspace{-1mm}
    \caption{The data statistics of four subsets in \textit{SKA-Bench}. `\#num P' and `\#num N' refer to the average number of positive units and noisy units. And `\#avg P token' denotes the average number of tokens in positive units. `\#data' refers to the numbers of instances in each subsets. The calculation of tokens is based on GPT-4o's tokenizer. `Expert Time' refers to the median time for each question spent on annotation by human experts.}\label{dataset}
\vspace{-1mm}
\end{table*}

\begin{figure}
\centering
\includegraphics[scale=0.50]{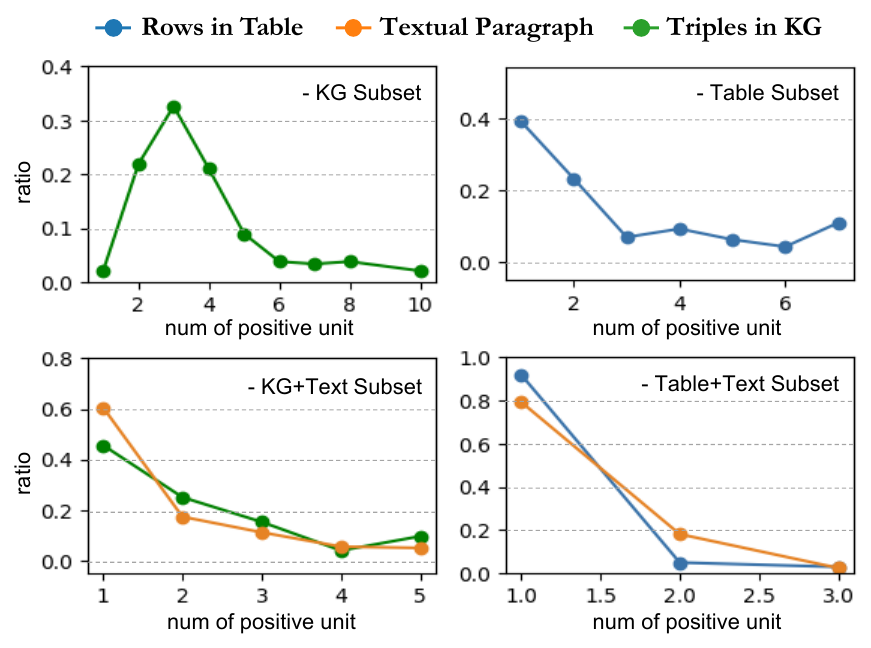}
\vspace{-2mm}
\caption{The distribution of the number of positive units across four \textit{SKA-Bench} subsets.} \label{dataset2}
\vspace{-2mm}
\end{figure}

\subsubsection{Noisy Units Construction}
For \texttt{KG} subset and \texttt{KG+Text} subset, we regard all knowledge units in the knowledge base except for the positive units as noisy units. The raw tables in the \texttt{Table} subset and \texttt{Table+}\texttt{Text} subset are typically short (average <17.9 rows), making it hard to comprehensively evaluate the table knowledge understanding of LLMs. Therefore, we introduce an automated noisy data synthesis process as follows.

First, we leverage LLMs with existing SKA-QA instances to generate noisy units. To ensure the diversity of synthesized units, we alternately use GPT-4o~\cite{gpt4} and DeepSeek-v3~\cite{deepseek-v3} during this process. Meanwhile, we also need to ensure that the synthesized noisy units do not affect the correctness of the answers. To achieve this, we prompt LLM (utilizing DeepSeek-v3) with QA and positive units to derive the ``\textit{conditions}'' that must be satisfied by the rows for answering the question. LLM then verifies whether the generated noisy units meet these ``\textit{conditions}''. If the response is ``\textit{Yes}'', the noisy units need to be re-generated by LLMs. After the noise synthesis process, three human experts conduct a manual review of \texttt{Table} subset and \texttt{Table+Text} subset to evaluate whether the synthetic noise is unsafe and affect the original answers. The review results show that the accuracy rate is 92.5\%, and erroneous noise has been deleted.

\subsection{Dataset Statistic}
Through the aforementioned construction pipeline, we have completed constructing \textit{SKA-Bench} instances as shown in Fig.~\ref{instance}, which consist of four main components: question, answer, positive knowledge units, and noisy knowledge units. Detailed statistics are presented in Table~\ref{dataset}. Additionally, we detail the human annotation results, i.e., the number of positive units across the four subsets, as shown in the Fig.~\ref{dataset2}.

\subsection{Testbeds Construction}
As shown in Fig.~\ref{intro}, inspired by~\citet{rgb} in text understanding evaluation, we construct the four testbeds based on \textit{SKA-Bench} instances to evaluate the following fundamental capabilities of LLMs in structured knowledge (SK) understanding: 

     \vspace{0.5mm}
    
    \noindent $\bullet$ \textbf{Noise Robustness.} 
    Here, we define noise as the remaining triples in the KG subgraph or the irrelevant rows in the table. We incorporate noise units of varying proportions into the positive knowledge units as the knowledge base to evaluate whether the LLM can robustly provide accurate answers. Considering the differences in the token counts across different knowledge units, we use the total token length as the split standard to construct test sets. Specifically, we construct four test sets \{1k, 4k, 12k, 24k\} for the \texttt{Table} and \texttt{KG} subsets, and three test sets \{4k, 12k, 24k\} for the \texttt{Table+Text} and \texttt{KG+Text} subsets, with the detailed statistics shown in Table~\ref{scale}. Additionally, to eliminate the influence of the knowledge unit order, we randomly shuffle the SK units in the KG and text units with a random seed of 42, while preserving the original order of the SK units in the Table. 

\begin{table}
\centering
\renewcommand{\arraystretch}{1.0}
\resizebox{0.47\textwidth}{!}{
\begin{tabular}{lccc}
\toprule
\textbf{Subset}                   & \textbf{\#num SK} & \textbf{\#num unSK} & \textbf{\#token}  \\ \midrule
\textbf{\textit{SKA-Bench}}-\texttt{KG-1k}          & 34.23    & -          & 637.64   \\
\textbf{\textit{SKA-Bench}}-\texttt{KG-4k}          & 150.34   & -          & 2831.40  \\
\textbf{\textit{SKA-Bench}}-\texttt{KG-12k}         & 604.35   & -          & 11394.18 \\
\textbf{\textit{SKA-Bench}}-\texttt{KG-24k}         & 1167.19  & -          & 22036.82 \\ \midrule
\textbf{\textit{SKA-Bench}}-\texttt{Table-1k}       & 29.39    & -          & 777.45   \\
\textbf{\textit{SKA-Bench}}-\texttt{Table-4k}       & 130.39   & -          & 3268.45  \\
\textbf{\textit{SKA-Bench}}-\texttt{Table-12k}      & 488.00   & -          & 12054.15 \\
\textbf{\textit{SKA-Bench}}-\texttt{Table-24k}      & 958.78   & -          & 23595.51 \\ \midrule
\textbf{\textit{SKA-Bench}}-\texttt{KG+Text-4k}     & 11.37    & 2.91       & 3172.54  \\
\textbf{\textit{SKA-Bench}}-\texttt{KG+Text-12k}    & 40.84    & 6.79       & 7417.67  \\
\textbf{\textit{SKA-Bench}}-\texttt{KG+Text-24k}    & 153.45   & 19.11      & 21644.20 \\ \midrule
\textbf{\textit{SKA-Bench}}-\texttt{Table+Text-4k}  & 25.82    & 14.58      & 3510.74  \\
\textbf{\textit{SKA-Bench}}-\texttt{Table+Text-12k} & 75.81    & 119.01     & 11899.54 \\
\textbf{\textit{SKA-Bench}}-\texttt{Table+Text-24k} & 165.81   & 369.01     & 23070.81 \\ \bottomrule
\end{tabular}}
\vspace{-1mm}
\caption{The data statistics for subsets with different scales of structured knowledge (SK) bases. `\#num SK' represents the number of structured knowledge units, `\#num unSK' represents the number of unstructured knowledge units in hybrid subsets. And `\#token' represents the total number of tokens in the knowledge bases.}
\vspace{-1mm}
\label{scale}
\end{table}

\begin{table*}[ht]
\centering
\renewcommand{\arraystretch}{1.0}
\resizebox{\textwidth}{!}{
\begin{tabular}{lcccccccccccccc}
\toprule
\multirow{2}{*}{\textbf{Model}}& \multicolumn{4}{c}{\textbf{\texttt{KG}}}       & \multicolumn{4}{c}{\textbf{\texttt{Table}}}& \multicolumn{3}{c}{\textbf{\texttt{KG+Text}}}& \multicolumn{3}{c}{\textbf{\texttt{Table+Text}}}\\ \cmidrule(lr){2-5}\cmidrule(lr){6-9}\cmidrule(lr){10-12}\cmidrule(lr){13-15}
                       & \texttt{\;\;\;\,1k\,\;\,\;}   & \texttt{\;\;\;\,4k\,\;\,\;}   & \texttt{\;\,\;12k\;\,\,}  & \texttt{\;\,\;24k\;\,\,}  & \texttt{\;\;\;\,1k\,\;\,\;}   & \texttt{\;\;\;\,4k\,\;\,\;}   & \texttt{\;\;\,12k\;\,\,}  & \texttt{\;\;\,24k\;\,\,}  & \texttt{\;\;\;\,4k\,\;\,\;}   & \texttt{\;\;\,12k\;\,\,}  & \texttt{\;\,\;24k\;\,\,}  & \texttt{\;\;\,\;4k\,\;\,\;}   & \texttt{\;\,\;12k\;\,\,}  & \texttt{\;\,\;24k\;\,\,}  \\ \midrule
\multicolumn{15}{c}{\textit{Open Source LLMs}}       \\ \midrule
Llama3.1-8B            & \cellcolor{blue0}67.53 & \cellcolor{blue1}58.19 & \cellcolor{blue2}45.86 & \cellcolor{blue3}42.34 & \cellcolor{blue0}\underline{27.56} & \cellcolor{blue1}23.52 & \cellcolor{blue2}\underline{22.16} & \cellcolor{blue3}13.05 & \cellcolor{blue0}67.02 & \cellcolor{blue1}58.89 & \cellcolor{blue2}49.28 &  \cellcolor{blue0}30.27 & \cellcolor{blue1}18.44 & \cellcolor{blue2}12.48    \\
TableGPT-2             & \cellcolor{blue0}\underline{78.93}& \cellcolor{blue1}\textbf{66.76}& \cellcolor{blue2}\textbf{53.14} & \cellcolor{blue3}\underline{48.49} &\cellcolor{blue0}24.40  & \cellcolor{blue1}\underline{24.05} & \cellcolor{blue2}20.09     & \cellcolor{blue3}16.02     & \cellcolor{blue0}64.84   & \cellcolor{blue1}55.16   & \cellcolor{blue2}46.92  & \cellcolor{blue0}\underline{35.91}  & \cellcolor{blue1}25.63   & \cellcolor{blue2}25.60   \\
Qwen2.5-7B             & \cellcolor{blue0}72.45 & \cellcolor{blue1}60.00 & \cellcolor{blue2}47.98 & \cellcolor{blue3}40.97 & \cellcolor{blue0}\textbf{36.69}   & \cellcolor{blue1}\textbf{32.04}  & \cellcolor{blue2}\textbf{30.45}  & \cellcolor{blue3}\textbf{28.68}  & \cellcolor{blue0}\textbf{76.51}   & \cellcolor{blue1}62.82   & \cellcolor{blue2}51.83 & \cellcolor{blue0}\textbf{38.49}  &\cellcolor{blue1}\textbf{36.00}   & \cellcolor{blue2}\underline{28.56}     \\
GLM4-9B                & \cellcolor{blue0}\textbf{82.95}& \cellcolor{blue1}\underline{66.04}& \cellcolor{blue2}\underline{52.75} & \cellcolor{blue3}\textbf{49.95} & \cellcolor{blue0}19.55  & \cellcolor{blue1}17.71 & \cellcolor{blue3}16.77   & \cellcolor{blue2}\underline{17.26}  & \cellcolor{blue0}\underline{75.39}  & \cellcolor{blue1}\underline{65.14}   & \cellcolor{blue2}\textbf{55.29}   & \cellcolor{blue1}32.13  & \cellcolor{blue0}\underline{33.65}  & \cellcolor{blue2}\textbf{30.13}    \\
Mistral-7B             & \cellcolor{blue1}59.04 & \cellcolor{blue0}60.34 & \cellcolor{blue2}47.98 & \cellcolor{blue3}45.20 & \cellcolor{blue1}17.67   & \cellcolor{blue0}18.11  & \cellcolor{blue2}16.91  & \cellcolor{blue3}16.19   & \cellcolor{blue0}69.37   & \cellcolor{blue1}\textbf{66.97}   & \cellcolor{blue2}\underline{53.54}   & \cellcolor{blue0}29.21  & \cellcolor{blue1}25.40   & \cellcolor{blue2}15.83   \\ \midrule
\multicolumn{15}{c}{\textit{Advanced General-Purpose LLMs}}       \\ \midrule
DeepSeek-v3            & \cellcolor{blue0}85.06 &\cellcolor{blue1}
 \underline{73.93}& \cellcolor{blue2}\underline{65.85}& \cellcolor{blue3}\underline{59.08}& \cellcolor{blue0}\underline{54.42} & \cellcolor{blue1}\underline{51.83} & \cellcolor{blue2}\underline{47.58}&  \cellcolor{blue3}\underline{45.57} & \cellcolor{blue0}77.12   & \cellcolor{blue1}\underline{74.96}  & \cellcolor{blue2}\underline{68.87}  & \cellcolor{blue0}55.64  & \cellcolor{blue1}\underline{53.61} & \cellcolor{blue2}48.55    \\
GPT-4o                 & \cellcolor{blue0}\underline{85.33}& \cellcolor{blue1}73.42 & \cellcolor{blue2}63.04 & \cellcolor{blue3}58.61 & \cellcolor{blue0}51.39 & \cellcolor{blue1}45.18   & \cellcolor{blue2}40.55  & \cellcolor{blue3}38.24  & \cellcolor{blue0}\underline{77.38}  & \cellcolor{blue1}73.53   & \cellcolor{blue2}67.39   & \cellcolor{blue0}\underline{56.52} & \cellcolor{blue1}53.28   & \cellcolor{blue2}\underline{51.97}        \\
DeepSeek-R1\;\,         & \cellcolor{blue0}\textbf{89.95}& \cellcolor{blue1}\textbf{81.58}& \cellcolor{blue2}\textbf{70.32}& \cellcolor{blue3}\textbf{64.67}& \cellcolor{blue0}\textbf{61.96} & \cellcolor{blue1}\textbf{61.88} & \cellcolor{blue2}\textbf{61.02}& \cellcolor{blue3}\textbf{58.24} & \cellcolor{blue0}\textbf{83.14}  & \cellcolor{blue1}\textbf{78.67}  & \cellcolor{blue2}\textbf{71.92}  & \cellcolor{blue0}\textbf{62.24} & \cellcolor{blue1}\textbf{57.62} & \cellcolor{blue2}\textbf{56.97}   \\ \bottomrule
\end{tabular}}
\vspace{-2mm}
\caption{Detailed results of noise robustness analysis. The best results are marked \textbf{bold} and the second-best results are \underline{underlined} in each column. Cells with darker colors indicate the better performance under this subset.}\label{noise}
\vspace{-1mm}
\end{table*}

    
     \vspace{0.5mm}
    \noindent$\bullet$ \textbf{Order Insensitivity.} SK representation naturally does not depend on any specific order. And in retrieval-augmented scenarios~\cite{rag-survey}, the order of retrieved knowledge units tends to be disrupted. Therefore, we expect LLMs to be order-insensitive when understanding SK and capturing the semantic relationships between SK units. In this testbed, we provide SK bases with different permutations of SK units to test whether the LLM is sensitive to order. For SK units in KG and textual units, we position the positive knowledge units at the beginning, randomized positions, and the end of the knowledge base, denoting them as \{\textit{prefix, random, suffix}\}. For SK units in Table, we additionally introduce the original table order, denoted as \{\textit{original}, \textit{prefix}, \textit{random}, \textit{suffix}\}. Furthermore, we standardize the test sets to a scale of 4k tokens for \texttt{Table} and \texttt{KG} subsets, and 12k for \texttt{Table+Text} and \texttt{KG+Text} subsets.
    
     \vspace{0.5mm}
    \noindent$\bullet$ \textbf{Information Integration.} This ability requires LLMs to integrate multiple knowledge units to answer questions, including the integration of multiple SK units and the integration of heterogeneous data (SK+Text) units. Therefore, this testbed focuses on analyzing the performance of LLMs under these two settings. Specifically, we divide our dataset based on the number of knowledge units required to answer each question \{2, 3, 4, more than 4\} to evaluate the information integration capability of LLMs. Regarding dataset scale and order, we standardize the test set to a scale of 4k tokens for the \texttt{Table} and \texttt{KG} subsets, and 12k tokens for \texttt{Table+Text} and \texttt{KG+Text} subsets. Meanwhile, we randomly shuffle (with random seed 42) the SK units in \texttt{KG} and text units while preserving the original order of SK units in the \texttt{Table} subset.

\begin{figure}
\centering
\includegraphics[scale=0.57]{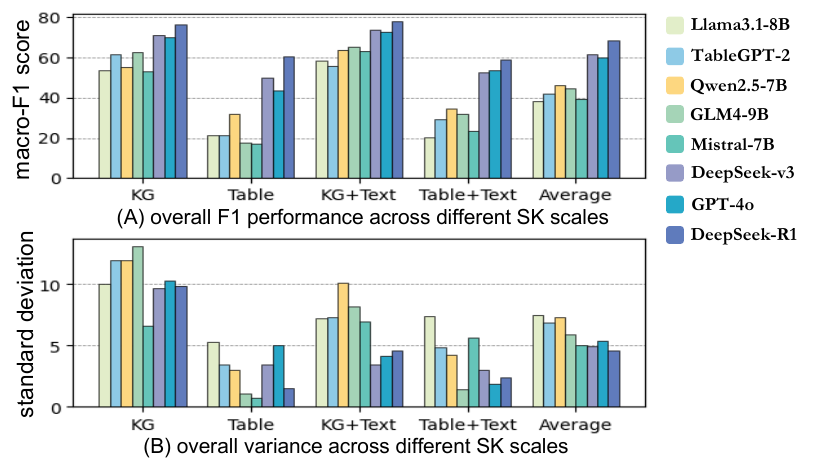}
\vspace{-6mm}
\caption{Overall noise robustness results on four subsets. `Average' represents the average results across on all results of four subsets. } \label{result1}
\vspace{-1mm}
\end{figure}

     \vspace{0.5mm}
    \noindent$\bullet$ \textbf{Negative Rejection.} We hope LLMs should minimize the occurrence of hallucination phenomena~\cite{hallucination} as much as possible when understanding SK. To evaluate this, we construct a negative rejection testbed, where the input SK base consists solely of noisy knowledge units. In this scenario, the LLMs are expected to respond with ``\textit{I don't know}'' or other rejection signals. In this testbed, the provided SK don't contain any positive units, ensuring broken reasoning paths to evaluate the refusal capability of LLMs. The dataset size and the ordering of knowledge units follow the same settings as ``\textit{Information Integration}'' testbed.  






\begin{table*}
\centering
\renewcommand{\arraystretch}{1.0}
\resizebox{\textwidth}{!}{
\begin{tabular}{lcccccccccccccc}
\toprule
\multirow{2}{*}{\textbf{Model}}& \multicolumn{3}{c}{\textbf{\texttt{KG}}}       & \multicolumn{4}{c}{\textbf{\texttt{Table}}}& \multicolumn{3}{c}{\textbf{\texttt{KG+Text}}}& \multicolumn{4}{c}{\textbf{\texttt{Table+Text}}}\\ \cmidrule(lr){2-4}\cmidrule(lr){5-8}\cmidrule(lr){9-11}\cmidrule(lr){12-15}
                       & \textit{prefix}   & \textit{random}   & \textit{suffix}  & \textit{original}  & \textit{prefix}   & \textit{random}   & \textit{suffix}  & \textit{prefix}  & \textit{random}   & \textit{suffix}  & \textit{original}  & \textit{prefix}   & \textit{random}  & \textit{suffix}  \\ \midrule
\multicolumn{15}{c}{\textit{Open Source LLMs}}       \\ \midrule
Llama3.1-8B            & \cellcolor{blue2}55.07  & \cellcolor{blue1}58.19 & \cellcolor{blue0}65.85 & \cellcolor{blue1}23.52 & \cellcolor{blue2}22.71     & \cellcolor{blue3}\underline{19.47}     & \cellcolor{blue0}\underline{24.57}     & \cellcolor{blue1}61.85     & \cellcolor{blue2}58.89    & \cellcolor{blue0}62.55   & \cellcolor{blue3}18.44     & \cellcolor{blue1}22.37     & \cellcolor{blue2}18.53   & \cellcolor{blue0}24.41    \\
TableGPT-2             & \cellcolor{blue0}\textbf{82.07} & \cellcolor{blue2}\textbf{66.76} & \cellcolor{blue1}\underline{77.36} & \cellcolor{blue1}\underline{24.05}    & \cellcolor{blue0}\underline{26.40}     & \cellcolor{blue3}17.25     & \cellcolor{blue2}21.62     & \cellcolor{blue0}57.47     & \cellcolor{blue1}55.16   & \cellcolor{blue2}54.53   & \cellcolor{blue2}25.63   & \cellcolor{blue0}36.75   & \cellcolor{blue3}\underline{24.44}      & \cellcolor{blue1}28.03    \\
Qwen2.5-7B             & \cellcolor{blue0}78.60 & \cellcolor{blue2}60.00 & \cellcolor{blue1}75.70 & \cellcolor{blue1}\textbf{32.04}  & \cellcolor{blue0}\textbf{33.26}  & \cellcolor{blue3}\textbf{24.07}  & \cellcolor{blue2}\textbf{31.38}  & \cellcolor{blue1}\underline{64.89}     & \cellcolor{blue2}62.82   & \cellcolor{blue0}\underline{67.74}   & \cellcolor{blue2}\textbf{36.00}  & \cellcolor{blue0}\textbf{48.29}  & \cellcolor{blue3}\textbf{29.37}    & \cellcolor{blue1}\textbf{37.46}          \\
GLM4-9B                & \cellcolor{blue1}\underline{81.30} & \cellcolor{blue2}\underline{66.04} & \cellcolor{blue0}\textbf{82.55} & \cellcolor{blue1}17.71 & \cellcolor{blue0}21.15  & \cellcolor{blue3}12.38  & \cellcolor{blue2}16.22  & \cellcolor{blue0}\textbf{70.05}     & \cellcolor{blue2}\underline{65.14}   & \cellcolor{blue1}\textbf{69.34}   & \cellcolor{blue1}\underline{33.65}  & \cellcolor{blue0}\underline{41.20}   & \cellcolor{blue3}23.89   & \cellcolor{blue2}\underline{31.14} \\
Mistral-7B             & \cellcolor{blue0}73.28  & \cellcolor{blue2}60.34 & \cellcolor{blue1}64.30   & \cellcolor{blue1}18.11  & \cellcolor{blue0}21.32  & \cellcolor{blue3}14.76  & \cellcolor{blue2}15.92   & \cellcolor{blue2}63.19     & \cellcolor{blue0}\textbf{66.97}   & \cellcolor{blue1}66.36   & \cellcolor{blue2}25.40  & \cellcolor{blue0}33.16   & \cellcolor{blue3}15.44   & \cellcolor{blue1}27.78 \\ \midrule
\multicolumn{15}{c}{\textit{Advanced General-Purpose LLMs}}       \\ \midrule
DeepSeek-v3            & \cellcolor{blue1}\underline{84.40}& \cellcolor{blue2}\underline{73.93}& \cellcolor{blue0}\underline{87.52}& \cellcolor{blue0}\underline{51.83} & \cellcolor{blue2}\underline{49.32} & \cellcolor{blue3}\underline{44.75} & \cellcolor{blue1}\underline{51.31} & \cellcolor{blue0}\underline{76.81}    & \cellcolor{blue2}\underline{74.96}  & \cellcolor{blue1}\underline{76.41}  & \cellcolor{blue1}\underline{53.61}& \cellcolor{blue0}\underline{55.80}  & \cellcolor{blue3}47.02   & \cellcolor{blue2}49.06  \\
GPT-4o                 & \cellcolor{blue1}81.75 & \cellcolor{blue2}73.42 & \cellcolor{blue0}83.69 & \cellcolor{blue1}45.18  & \cellcolor{blue0}45.62  & \cellcolor{blue3}40.47  & \cellcolor{blue2}43.33  & \cellcolor{blue1}74.88    & \cellcolor{blue2}73.53   & \cellcolor{blue0}74.98   & \cellcolor{blue1}53.28  & \cellcolor{blue0}54.88    & \cellcolor{blue3}\underline{47.72}  & \cellcolor{blue2}\underline{52.23}  \\
DeepSeek-R1\;\,           & \cellcolor{blue0}\textbf{89.90}& \cellcolor{blue2}\textbf{81.58}& \cellcolor{blue1}\textbf{89.40}  & \cellcolor{blue2}\textbf{61.88} & \cellcolor{blue0}\textbf{67.11} & \cellcolor{blue3}\textbf{61.63}& \cellcolor{blue1}\textbf{64.36}  & \cellcolor{blue1}\textbf{79.60} & \cellcolor{blue2}\textbf{78.67} & \cellcolor{blue0}\textbf{81.12}& \cellcolor{blue2}\textbf{57.62}  & \cellcolor{blue0}\textbf{59.28} & \cellcolor{blue3}\textbf{53.04} & \cellcolor{blue1}\textbf{57.97} \\ \bottomrule
\end{tabular}}
\vspace{-2mm}
\caption{Results of order insensitivity analysis. The best results are marked \textbf{bold} and the second-best results are \underline{underlined} in each column. Cells with darker colors indicate the better performance under this subset.}\label{sequence}
\vspace{-1mm}
\end{table*}
\begin{figure}
\centering
\includegraphics[scale=0.24]{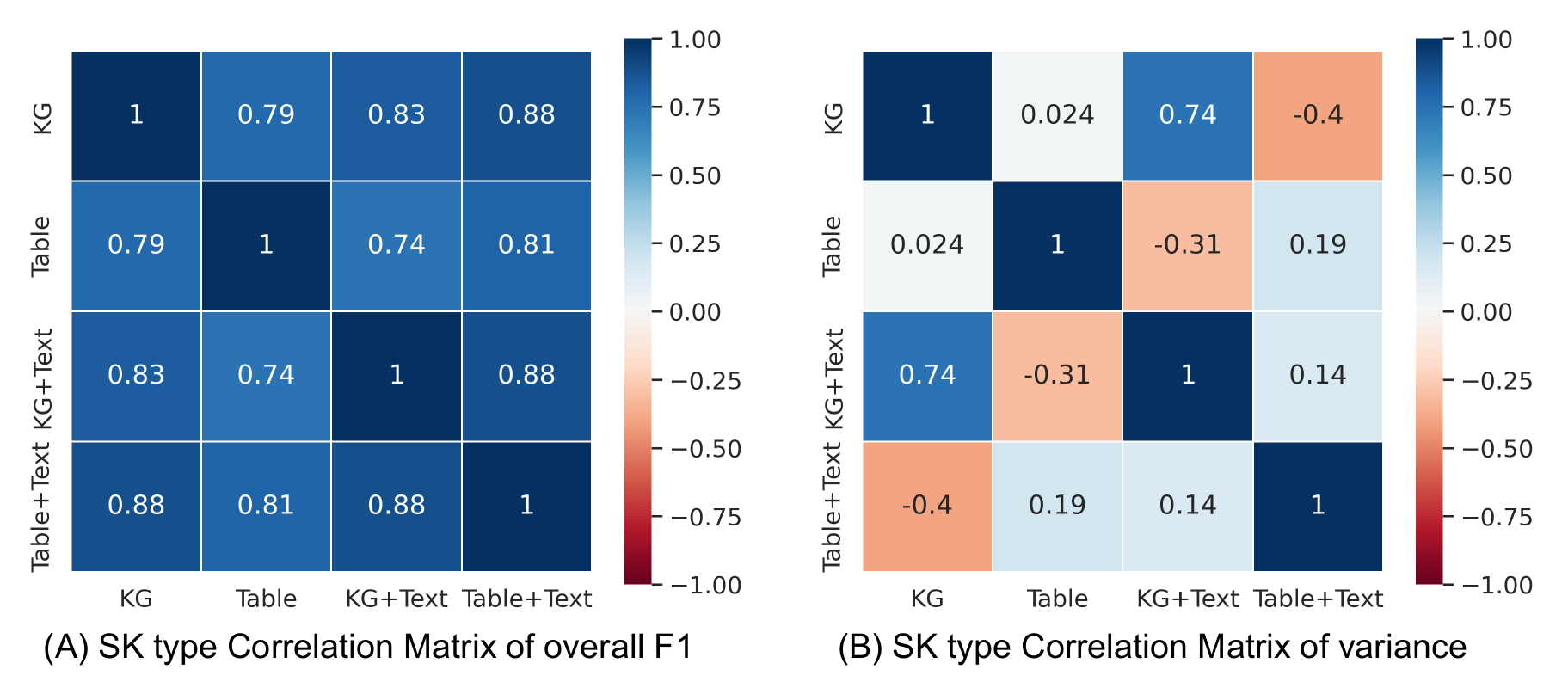}
\vspace{-2mm}
\caption{Correlation coefficients of overall F1 and standard deviation across 4 SK types under noise robustness testbed. } \label{result1-1}
\vspace{-1mm}
\end{figure}

\section{Experiments}
\subsection{Experimental Settings}
\paragraph{Models.}
Our evaluation is based on popular large language models (LLMs) with a context window of at least 24k tokens. Our evaluated LLMs include advanced general-purpose LLMs: DeepSeek-v3~\cite{deepseek-v3}, GPT-4o~\cite{gpt4}, DeepSeek-R1~\cite{deepseekr1} and common open-source LLMs: Llama-3.1-8B-Instruct~\cite{llama3}, Qwen2.5-7B-Instruct~\cite{qwen2.5}, GLM4-9B-Chat~\cite{glm4}, and Mistral-7B-Instruct-v0.3~\cite{mistral}. Moreover, we also evaluate the table-specific open-source LLM TableGPT-2~\cite{tablegpt2}, which are trained based on Qwen2.5-7B.

\paragraph{Evaluation Metric.}
To evaluate \textit{SKA-Bench}, we utilize the macro-F1 score as our metrics, which measures the agreement between the predicted answer list and the gold answer list. For the negative rejection testbed, we adopt the ``Rejection Rate'' as the evaluation metric, which reflects the proportion of instances where the LLMs provide a refusal response out of the total number of test samples when only noisy knowledge units are provided.

\begin{figure}
\centering
\includegraphics[scale=0.58]{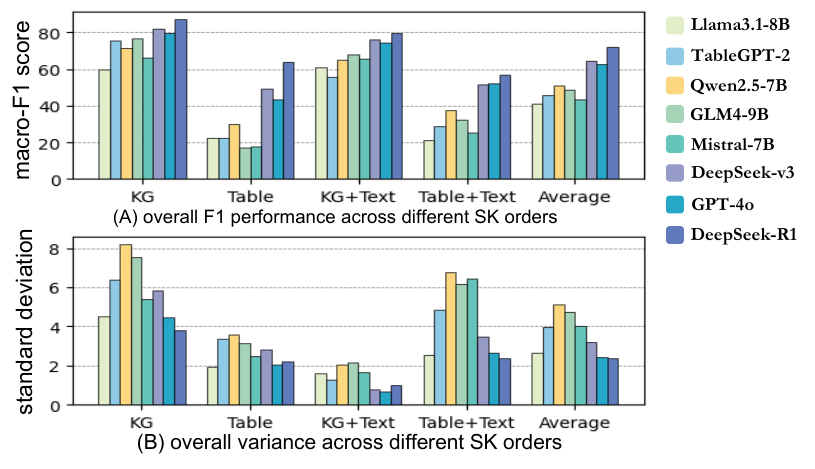}
\vspace{-6mm}
\caption{Overall order insensitivity results on four subsets. `Average' represents the average results across on all results of four subsets. } \label{result2}
\vspace{-1mm}
\end{figure}

\subsection{Noise Robustness Analysis}


From the results in Table~\ref{noise}, it can be observed that as 
the length of SK input to LLM increases,
the performance degradation across various LLMs becomes significantly pronounced. In particular, Llama3.1-8B exhibits a dramatic decline of up to 58.77\% when evaluated on the \texttt{Table+Text} subset from 4k to 24k scale. DeepSeek-R1 demonstrates optimal results across all subsets, whereas GLM4-9B and Qwen2.5-7B achieve relatively competitive performance among the smaller models with the 7-10B parameters.

To further analyze model performance on different data types, we present the mean and standard deviation of F1 scores, and their correlation matrix across 4 subsets, as shown in Fig.~\ref{result1} and~\ref{result1-1}. We can observe that the performance trends of different LLMs across 4 SK types are similar in general, with all spearman $\rho>0.64$. However, there are significant differences in the noise robustness of different LLMs across 4 SK types as shown in Fig.~\ref{result1-1}(B). GLM4-9B can perform well on the \texttt{KG} subset but struggles to understand Table data, and TableGPT-2 leverages large-scale table-related task instruction fine-tuning on the base model Qwen2.5-7B, but its performance on both the \texttt{Table} and \texttt{Table+Text} subsets is less satisfactory. We attribute this to the loss of generalization capabilities due to its specialized training, making it less adaptable to unseen table formats and other data modalities. Furthermore, we observe that DeepSeek-R1 achieves the lowest average standard deviation, exhibiting the strongest noise robustness. \textbf{This suggests that current LLMs are evolving towards greater robustness against noise.}

\begin{figure}
\centering
\includegraphics[scale=0.25]{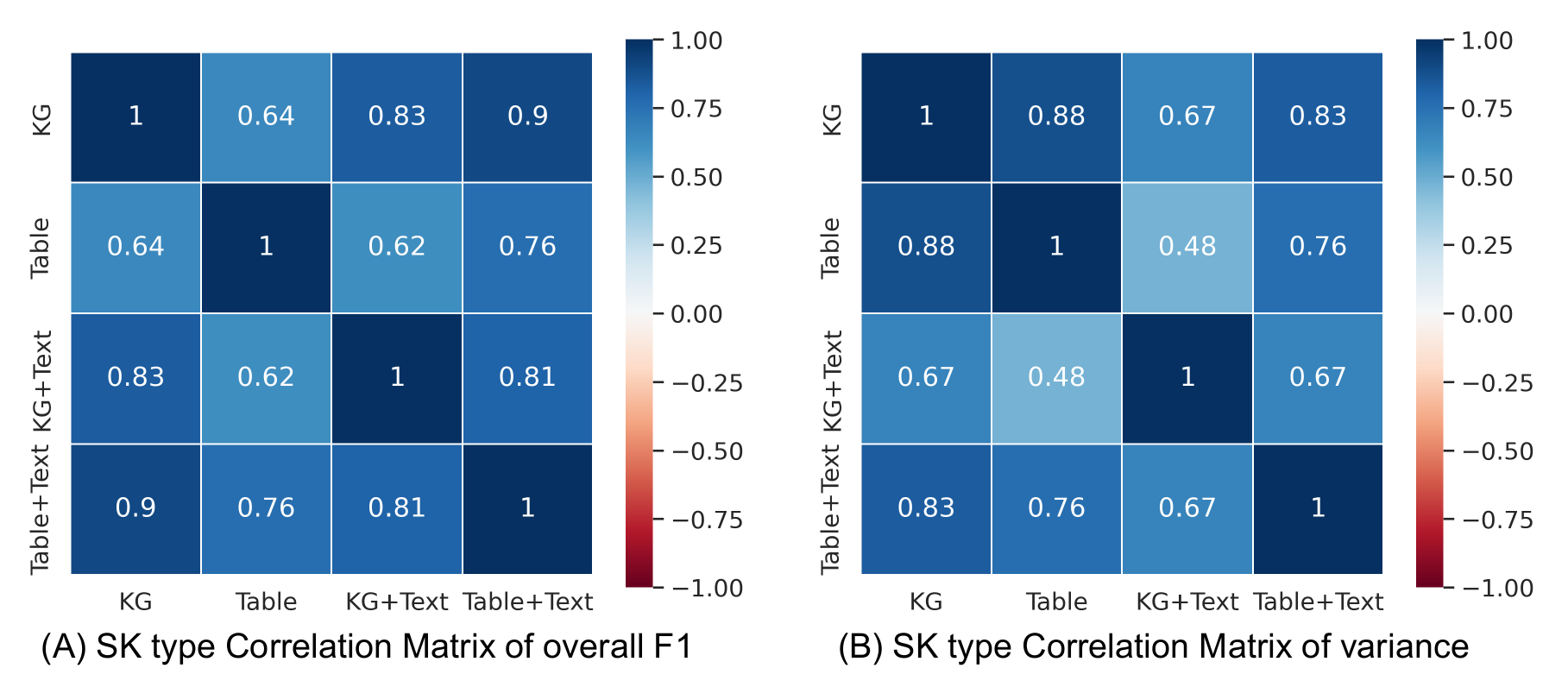}
\vspace{-6mm}
\caption{Correlation coefficients of overall F1 and standard deviation across 4 SK types in order insensitivity testbed. } \label{result2-1}
\vspace{-1mm}
\end{figure}

\subsection{Order Insensitivity Analysis}

\begin{figure*}
\centering
\includegraphics[scale=0.6]{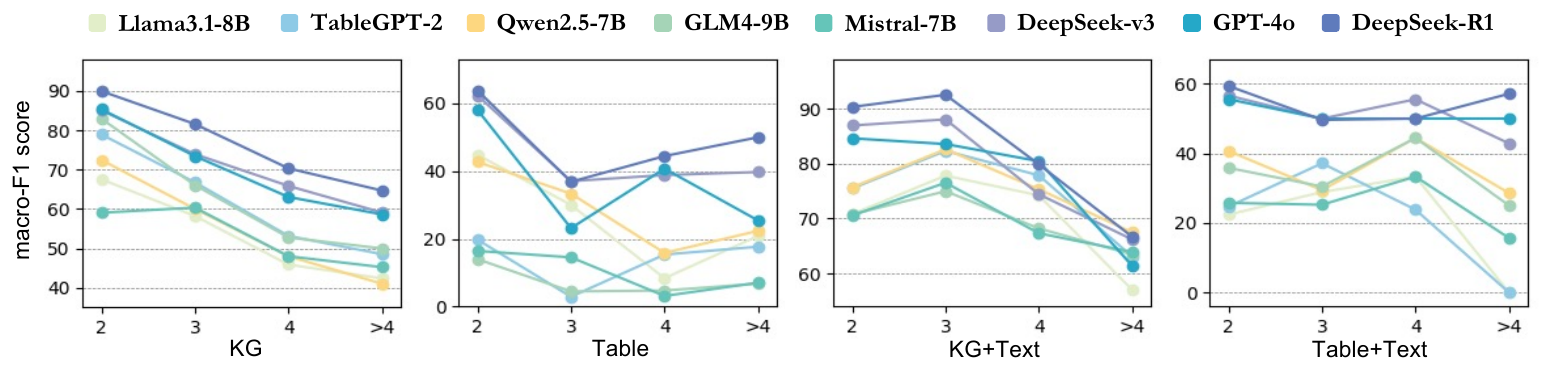}
\vspace{-3mm}
\caption{Information Integration results on four subsets demonstrates the variation in F1 score as the number of required positive units increases. } \label{result3}
\vspace{-3mm}
\end{figure*}

From the results in Table~\ref{sequence}, we can observe that when the positive units are concentrated in the prefix or suffix of the structured knowledge base, models tend to focus on them more effectively and achieve better response performance. However, when the positive units are randomly scattered throughout the knowledge base, LLMs often experience the ``Lost in the Middle''~\cite{lostinthemiddle} phenomenon, making them more likely to respond incorrectly. This suggests that for structured knowledge retrieval scenarios, \textbf{recalling positive units as early as possible can effectively enable LLMs to focus on them, thereby improving performance.}  

In Fig.~\ref{result2} and~\ref{result2-1}, we present the mean and standard deviation of F1 scores, and their correlation matrix across different subsets under the order insensitivity testbed. As illustrated in Fig.~\ref{result2-1}, we can observe that the order sensitivity of LLMs across 4 SK types exhibits a positive correlation, and so does their F1 performance. From the perspective of standard deviation, models that are insensitive to the order of SK are generally either weaker-performing LLMs, such as Llama3.1-8B, or exceptionally strong-performing LLMs, such as DeepSeek-R1. The former consistently exhibits weaker capabilities across various order settings, while the latter demonstrates stronger understanding and reasoning abilities, \textbf{suggesting that current LLMs are evolving towards greater robustness and less sensitive to the order of knowledge units.}

\begin{table}
\centering
\renewcommand{\arraystretch}{1.0}
\resizebox{0.46\textwidth}{!}{
\begin{tabular}{lcccccc} \\ \toprule
\textbf{Model}       & \textbf{\texttt{KG}}    & \textbf{\texttt{Table}} & \textbf{\texttt{KG+Text}} & \textbf{\texttt{Table+Text}} & \textbf{Avg.}  \\ \midrule
\multicolumn{6}{c}{\textit{Open Source LLMs}}       \\ \midrule
Llama3.1-8B & 49.36 & 47.46 & 48.21   & 56.57      & 50.40 \\
TableGPT-2  & \textbf{83.69} & \textbf{70.85} & \textbf{85.13}   & \textbf{93.94}      & \textbf{83.40} \\
Qwen2.5-7B  & \underline{81.55} & \underline{70.17} & \underline{75.90}   & \underline{80.81}      & \underline{77.11} \\
GLM4-9B     & 69.96 & 61.69 & 63.59   & 71.72      & 66.74 \\
Mistral-7B  & 61.37 & 62.71 & 51.28   & 53.03      & 57.10 \\ \midrule
\multicolumn{6}{c}{\textit{Advanced General-Purpose LLMs}}       \\ \midrule
DeepSeek-v3 & 78.54 & 69.83 & 58.97   & 69.70      & 69.26 \\
GPT-4o      & \underline{87.98} & \textbf{73.56} & \textbf{76.92}   & \underline{80.81}      & \textbf{79.82} \\
DeepSeek-R1 & \textbf{91.42} & \underline{72.88} & \underline{68.21}   & \textbf{82.32}           & \underline{78.71} \\ \bottomrule
\end{tabular}}
\vspace{0mm}
\caption{Negative Rejection results on four subsets.}
\vspace{-1mm}
\label{rejection}
\end{table}

\subsection{Information Integration Analysis}

From the results shown in Fig.~\ref{result3}, it can be observed that as the number of knowledge units required increases, the overall performance of the LLMs tends to decline. This phenomenon is more pronounced in the \texttt{KG} and \texttt{KG+Text} subsets. We believe this is due to the fact that noisy knowledge units and positive knowledge units in the KG are derived from subgraph. Many noisy units share the same entities or relations as the positive units and exhibit higher semantic similarity, which more significantly impacts the LLM's understanding. In contrast, the row units of table data are relatively more semantically independent, so this downward trend is less noticeable in the \texttt{Table} subset.

In terms of understanding heterogeneous data, it is evident that as the volume of heterogeneous data increases, the performance of most LLMs declines quite substantially. Notably, in the \texttt{Table+Text} subset with >4 heterogeneous units, the advanced LLMs such as DeepSeek-R1 and GPT-4o still maintain relatively strong performance, whereas smaller LLMs like TableGPT-2 and Llama3.1-8B struggle to generate correct answers. \textbf{Thus, we consider enhancing the ability of smaller LLMs to understand heterogeneous data to be a promising research direction worthy of further exploration.}

\subsection{Negative Rejection Analysis}

The results in Table~\ref{rejection} present the rejection rates when only noisy knowledge units are provided. Overall, there is a certain positive correlation between the structured knowledge understanding performance of the LLMs and its negative rejection ability. However, we find that even DeepSeek-R1, with a negative rejection rate of 78.71\%, remains vulnerable to noise interference. To our surprise, compared to Qwen2.5-7B, TableGPT-2 after fine-tuning with table-specific instructions, demonstrates stronger negative rejection ability, even surpassing GPT-4o and DeepSeek-R1. \textbf{Therefore, how to strike a balance between improving the LLM's performance and enhancing its negative rejection ability remains challenging.}

\section{Conclusion}
In this paper, we introduce a fine-grained structured knowledge (SK) understanding benchmark, \textbf{\textit{SKA-Bench}}, designed to provide a more comprehensive and rigorous evaluation for LLMs in understanding SK. The instances in \textit{SKA-Bench} consist of a question, an answer, positive knowledge units, and noisy knowledge units, offering greater flexibility and scalability. Through varying the order and scale of knowledge units within the knowledge base, we construct four specialized testbeds to evaluate key capabilities: \textit{Noise Robustness}, \textit{Order Insensitivity}, \textit{Information Integration}, and \textit{Negative Rejection}. Empirical results demonstrate that even powerful LLMs like GPT-4o and DeepSeek-R1 still lack comprehensive understanding and reasoning capabilities for SK. Their performance is significantly influenced by factors such as the amount of noise, order of knowledge units, and hallucinations.  

\section*{Limitations}
Although \textit{SKA-Bench} offers a more comprehensive and rigorous benchmark for evaluating structured knowledge understanding of LLMs, certain limitations warrant careful consideration, as summarized below. 
(1) \textit{SKA-Bench} is limited to English only and does not yet capture the performances of LLMs in understanding structured knowledge across multiple languages. (2) Constrained by resource limitations, although our \textit{SKA-Bench} instances have the capability to construct longer structured knowledge bases (even >64k tokens), we have not yet explored the performance of LLMs at this scale. (3) Potential biases introduced by dataset selection are inevitable. We have made effort to minimize this impact through careful dataset selection and evaluation design. For dataset selection, {\textit{SKA-Bench}} adopts widely used and well-recognized benchmark datasets for each respective data type (e.g., {\textit{WebQSP}}, \textit{{CWQ}}, \textit{{WTQ}}, \textit{{TableBench}}, \textit{{StarK-Amazon}}, and \textit{{HybridQA}}). These datasets focus on general domains and do not require extensive domain-specific knowledge, which helps to mitigate the bias introduced by domain dependence. For evaluation design, our primary focus is on whether LLMs can comprehensively understand various types of structured knowledge, rather than investigating the ``performance differences between different modalities''. (4) Beyond its role as a structured knowledge understanding dataset, \textit{SKA-Bench} can be effectively extended to various task scenarios, such as Text2Query (i.e., Text2SQL~\cite{text2sql} and Text2SPARQL~\cite{text2sparql}), structured knowledge retrieval~\cite{skretrieve}, and evaluations of knowledge-augmented systems~\cite{kag,pikerag}. These potential evaluation directions remain unexplored. 

\section*{Ethics Statement}
In this paper, we construct \textit{SKA-Bench}, which is expanded and modified based on the existing 6 structured knowledge understanding evaluation datasets. These datasets have stated that there are no ethical concerns. Moreover, we also incorporate manual annotation and manual synthetic data verification to ensure that it does not violate any ethics.

\section*{Acknowledgments}
This work is founded by National Natural Science Foundation of China (NSFC62306276/NSFCU23B2055/NSFCU19B2027), Zhejiang Provincial Natural Science Foundation of China (No. LQ23F020017), Yongjiang Talent Introduction Programme (2022A-238-G), and Fundamental Research Funds for the Central Universities (226-2023-00138). This work was supported by Ant Group.

\bibliography{acl}


\appendix
\section{Original Datasets Details}
We provide a brief description of all the original structured knowledge understanding datasets we used and licenses below:

\begin{itemize}
\item \textbf{\textit{WebQSP}}~\cite{webqsp}. \webqsp (\textit{WebQSP}) is a semantic parse-based KBQA dataset with 4,737 questions coupled with SPARQL queries for KB question answering. The answers can be extracted through executing SPARQL queries on Freebase. The dataset is released under the Microsoft Research Data License Agreement.
\item \textbf{\textit{CWQ}}~\cite{cwq}. \cwq (\textit{CWQ}) is created on top of \textit{WebQSP} dataset with the intention of generating more complex (by incorporating compositions, conjunctions, superlatives or comparatives) questions in natural language. It consists of 34,689 examples, divided into 27,734 train, 3,480 dev, 3,475 test. And test set in original \textit{CWQ} dataset does not contain ``answer''. The whole software is licensed under the full GPL v2+. 
\item \textbf{\textit{WTQ}}~\cite{wtq}. \wtq (\textit{WTQ}) is a widely used table question answering (TableQA) dataset of 22,033 complex questions with average 2.14 hop on Wikipedia tables. The dataset is released under the Apache-2.0 license.
\item \textbf{\textit{TableBench}}~\cite{tablebench}. \textit{TableBench} is a comprehensive and complex benchmark, including 886 samples in 18 fields within four major categories of TableQA capabilities. The tables in \textit{TableBench} have an average of 6.68 columns and 16.71 rows, and the average reasoning steps of questions is 6.26. The dataset is released under the Apache-2.0 license.
\item \textbf{\textit{STaRK}}~\cite{stark}. \textit{STaRK} is a large-scale semi-structure retrieval benchmark on textual and relational knowledge bases, covering three domains. It consists of 263 human-generated questions and 33,627 synthesized questions. And this dataset is released under the MIT license.
\item \textbf{\textit{HybridQA}}~\cite{hybridqa}. \textit{HybridQA} is a question answering dataset based on heterogeneous knowledge, and each question is aligned with a Wikipedia table and multiple free-form corpora linked with the entities in the table. The questions are collected from crowd-workers, and designed to aggregate both table and text information, which means the lack of either form would render the question unanswerable. The dataset is released under the MIT license.
\end{itemize}

\begin{table}[ht]
    \centering
    \resizebox{0.46\textwidth}{!}{\begin{tabular}{lcccc}
        \toprule
        \textbf{Type} & \textbf{1-hop} & \textbf{2-hop} & \textbf{3-hop} & $\mathbf{\geq}$\textbf{4-hop} \\
        \midrule
        {\texttt{KG} subset}      & 7.59\%  & 22.07\%  & 16.26\%  & 54.09\%  \\
        {\texttt{KG+Text} subset} & 7.04\%    & 10.91\%     & 9.76\%     & 72.29\%    \\
        \bottomrule
    \end{tabular}}
    \caption{Hop count distributions in \texttt{KG} subset and \texttt{KG+Text} subset.}
    \label{hop}
\end{table}

\begin{table*}[ht]
    \centering
    \resizebox{0.8\textwidth}{!}{\begin{tabular}{lcccccccc}
        \toprule
        \multirow{2}{*}{\textbf{Model}} 
        &\multicolumn{4}{c}{\textbf{\texttt{KG}}} 
        &\multicolumn{4}{c}{\textbf{\texttt{KG+Text}}} \\ 
        \cmidrule(lr){2-5} \cmidrule(lr){6-9} 
        & Hard Noise & Mixed Noise & Easy Noise & \cellcolor{gray!20}Std. 
        & Hard Noise  & Mixed Noise  & Easy Noise  & \cellcolor{gray!20}Std. \\
        \midrule
        LLaMA3.1-8B    
        & 55.12 & 58.19 & 60.23 & \cellcolor{gray!20}2.10 
        & 65.68 & 67.02 & 66.96 & \cellcolor{gray!20}0.62 \\
        Qwen2.5-7B     
        & 58.36 & 60.00 & {67.82} & \cellcolor{gray!20}4.13 
        & 74.96 & 76.51 & {77.15} & \cellcolor{gray!20}0.92 \\
        DeepSeek-v3    
        & 71.09 & \textbf{73.93} & 74.28 & \cellcolor{gray!20}1.43 
        & \textbf{77.01} & 77.12 & 77.65 & \cellcolor{gray!20}\textbf{0.28} \\
        GPT-4o         
        & \textbf{72.10} & 73.42 & \textbf{74.82} & \cellcolor{gray!20}\textbf{1.11} 
        & 76.58 & \textbf{77.38} & \textbf{77.63} & \cellcolor{gray!20}0.45 \\
        \bottomrule
    \end{tabular}}
    \caption{Performance of four representative LLMs under different noise intensities on \texttt{KG} and \texttt{KG+Text} subsets.}
    \label{result:hop}
\end{table*}

\section{Effects of Noise Distribution in KG-related Subsets}
Due to the structural complexity of knowledge graphs, the degree of noise can vary across different triples. We hypothesize that hop count is closely related to semantic similarity, and this relationship may affect the noise distribution, thereby impacting model performance. To investigate this, we conduct an in-depth analysis of the hop count distribution in the existing \texttt{KG} subsets and \texttt{KG+Text} subsets, and provide experimental results to support this hypothesis. As shown in Table~\ref{hop}, we present the detailed hop count distributions for the \texttt{KG} and \texttt{KG+Text} subsets. It is worth noting that we report the actual distributions in \textit{SKA-Bench} rather than the original datasets, since some subgraphs were randomly truncated during the graph extraction process.

Thanks to the high flexibility of \textit{SKA-Bench}, we further present results under different noise intensities to more thoroughly evaluate the noise robustness of LLMs. In our experimental setup, the context size is consistently controlled at 4k, and knowledge units from the knowledge graph are arranged in a random order. Noise is categorized as ``Hard Noise'' ($\leq$2 hops), ``Mixed Noise'' (same as in main text), and ``Easy Noise'' ($\geq$4 hops). We report the performance of four representative LLMs, as shown in Table~\ref{result:hop}. The experimental results show that the variation in model performance is significantly correlated with the intensity of noise. Comparatively, stronger LLMs demonstrate greater robustness to noise, and this conclusion also holds when the noise level is expanded horizontally (as in the noise robustness testbed of the main text).

\section{Dataset Construction Details}
\begin{figure}
\centering
\includegraphics[scale=0.46]{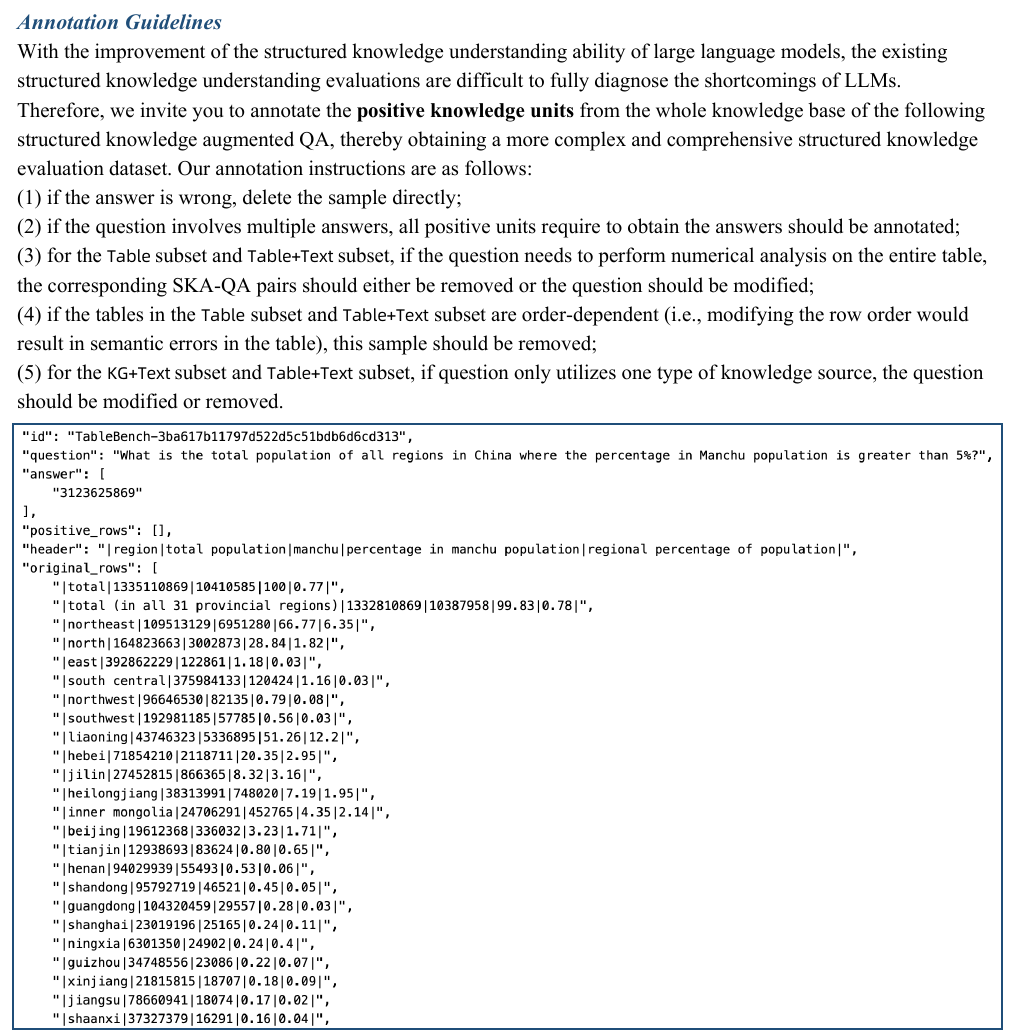}
\vspace{-1mm}
\caption{The annotation guidelines for annotators.} \label{guideline}
\end{figure}

The annotation guideline for ``Iterative Positive Units Annotation'' is shown in the Fig.~\ref{guideline}.

\begin{figure}[ht]
\centering
\includegraphics[scale=0.56]{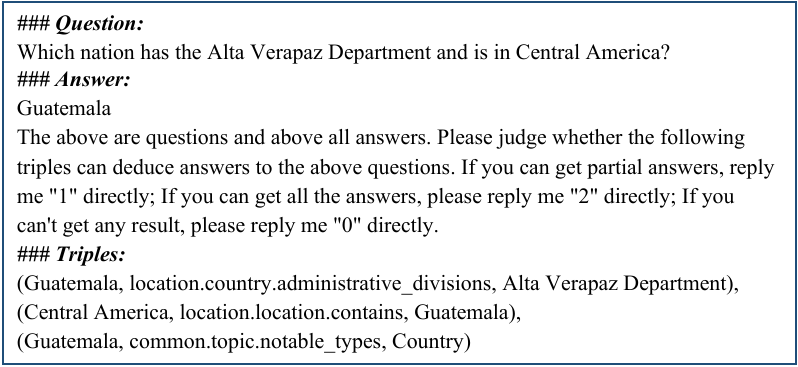}
\vspace{-1mm}
\caption{The prompt for checking Positive Units.} \label{pipeline1}
\end{figure}

\begin{figure}[ht]
\centering
\includegraphics[scale=0.56]{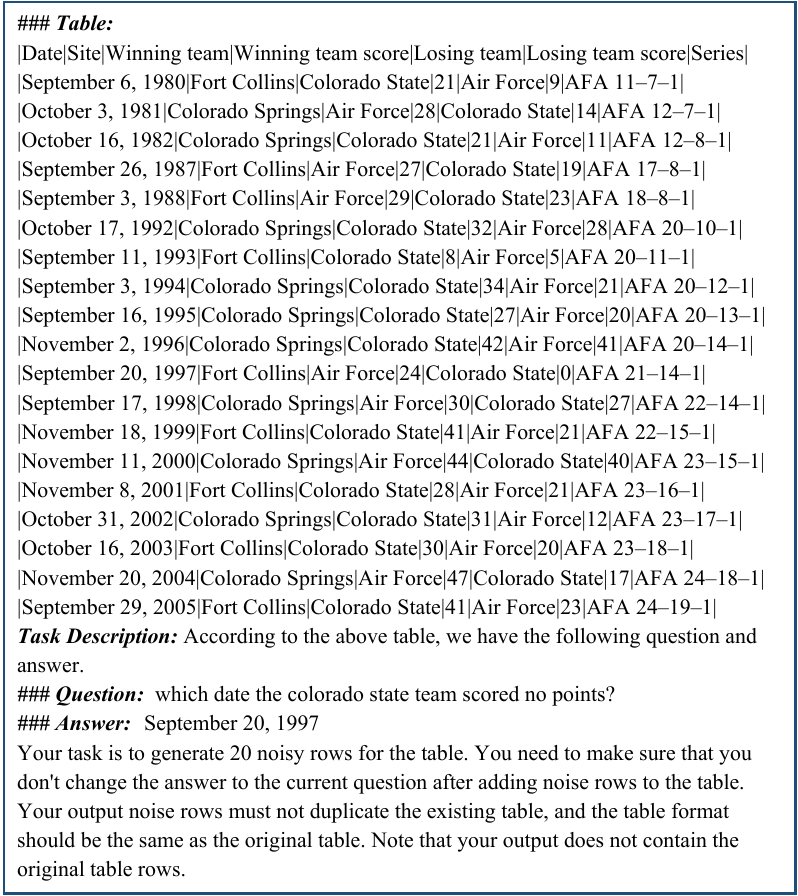}
\vspace{-1mm}
\caption{The prompt for Noisy Units synthesis.} \label{pipeline2}
\end{figure}

\begin{figure}[ht]
\centering
\includegraphics[scale=0.56]{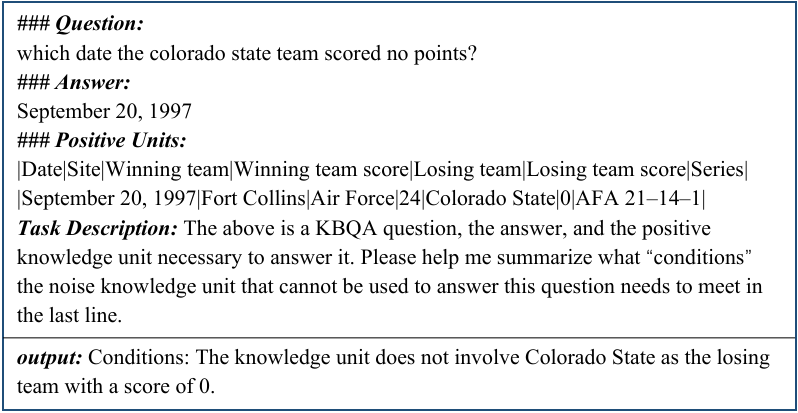}
\vspace{-1mm}
\caption{The prompt for ``contidition'' summarizing.} \label{pipeline3}
\end{figure}

\begin{figure}[ht]
\centering
\includegraphics[scale=0.56]{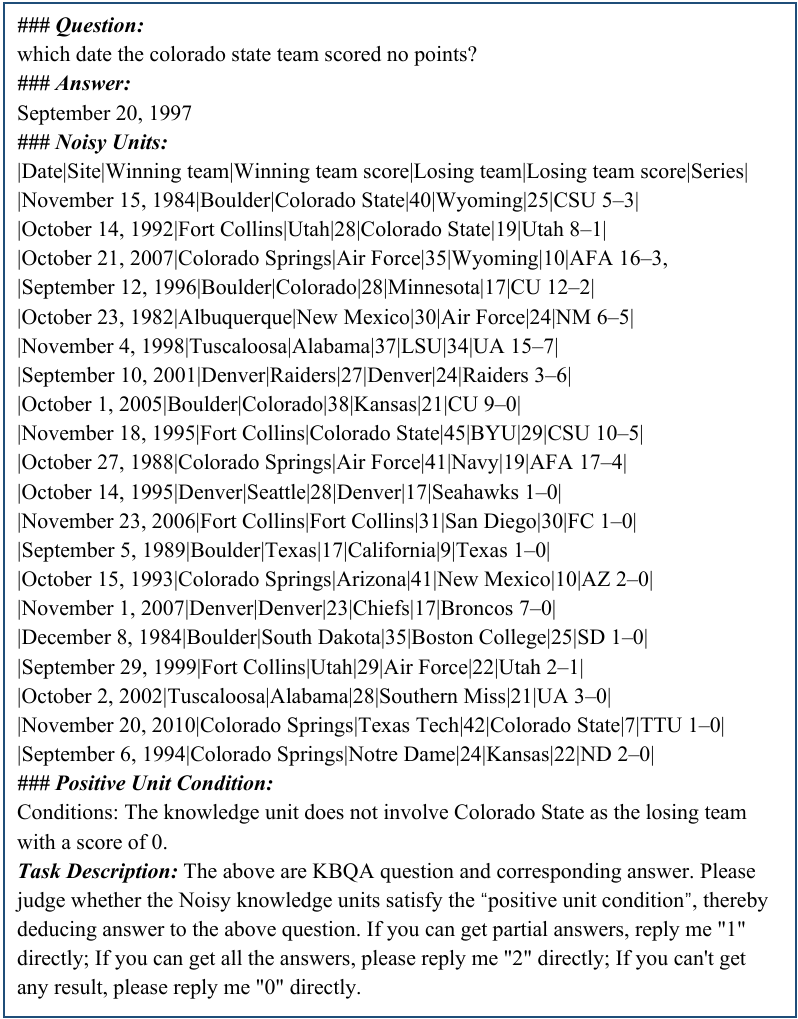}
\vspace{-1mm}
\caption{The prompt for checking Noisy Units.} \label{pipeline4}
\end{figure}

Moreover, we have presented specific examples of the part where LLMs are involved in the entire dataset construction process. Check satisfaction by LLM in Iterative Positive Units Annotation stage is shown in Fig.~\ref{pipeline1}. Noisy Synthesis process is shown in Fig.~\ref{pipeline2}. And ``condition'' of positive knowledge units summarizing and check satisfaction by LLMs in Noisy Units Construction stage are shown in Fig.~\ref{pipeline3} and Fig.~\ref{pipeline4}.

\section{Evaluation Prompt Template}
\begin{figure}[ht]
\centering
\includegraphics[scale=0.56]{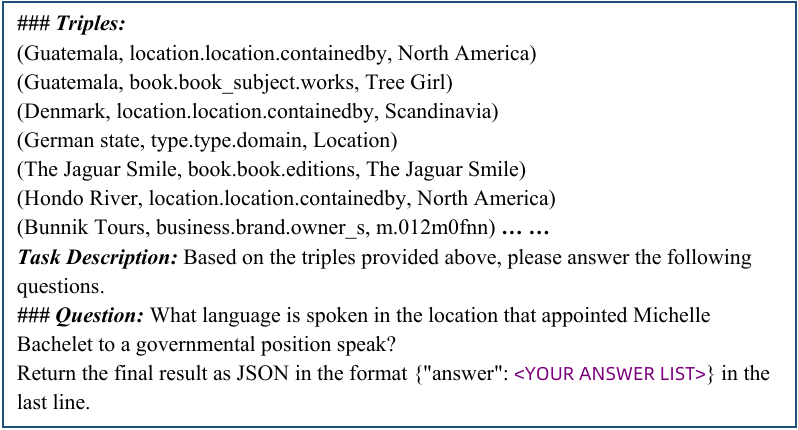}
\vspace{-1mm}
\caption{The prompt for \texttt{KG} subset in QA task.} \label{prompt1}
\end{figure}

\begin{figure}[ht]
\centering
\includegraphics[scale=0.56]{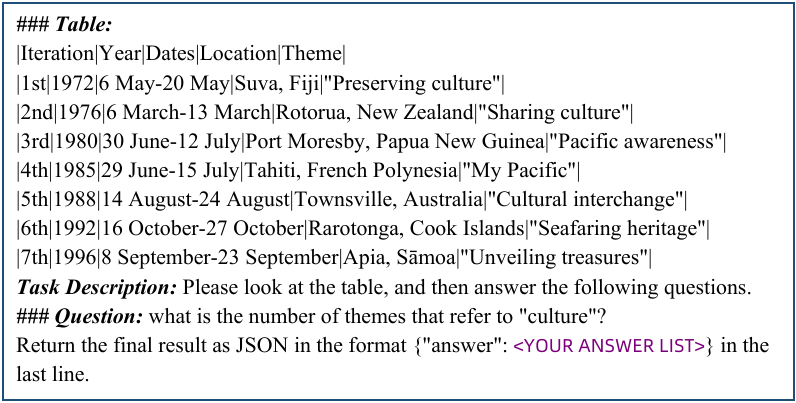}
\vspace{-1mm}
\caption{The prompt for \texttt{Table} subset in QA task.} \label{prompt2}
\end{figure}

\begin{figure}[ht]
\centering
\includegraphics[scale=0.56]{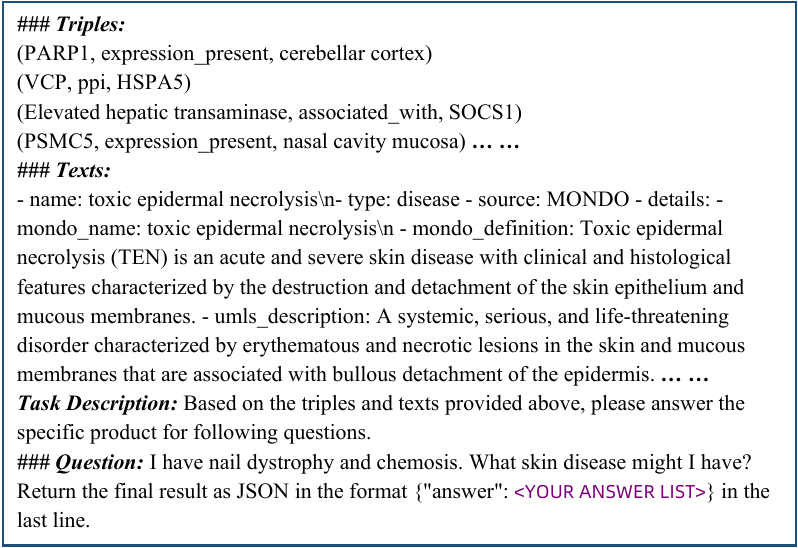}
\vspace{-1mm}
\caption{The prompt for \texttt{KG+Text} subset in QA task.} \label{prompt3}
\end{figure}

\begin{figure}[ht]
\centering
\includegraphics[scale=0.56]{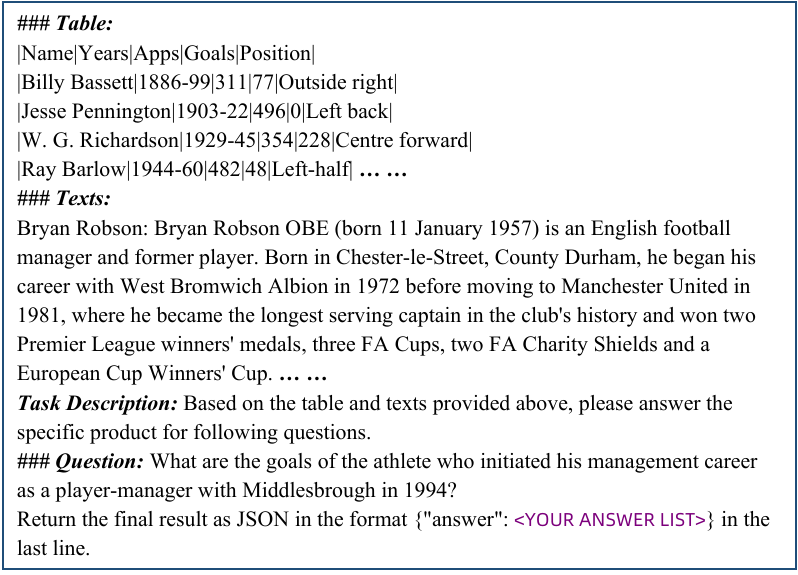}
\vspace{-1mm}
\caption{The prompt for \texttt{Table+Text} subset in QA task.} \label{prompt4}
\end{figure}

\begin{figure}[ht]
\centering
\includegraphics[scale=0.56]{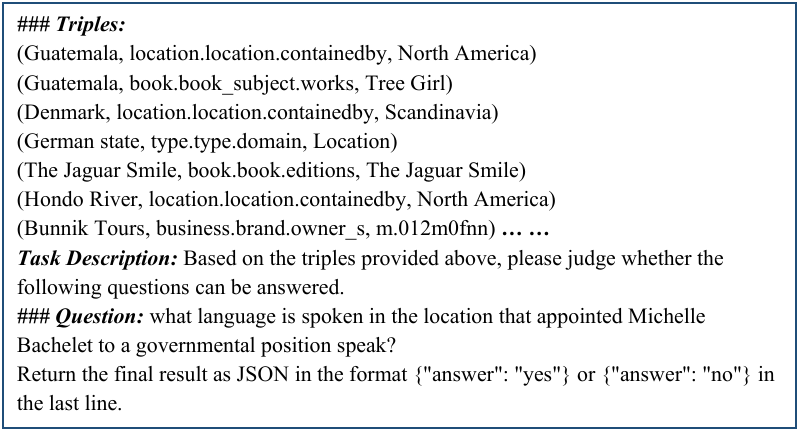}
\vspace{-1mm}
\caption{The prompt for \texttt{KG} subset in ``\textit{negative rejection}'' testbed.} \label{rejection1}
\vspace{-3mm}
\end{figure}

\begin{figure}[ht]
\centering
\includegraphics[scale=0.56]{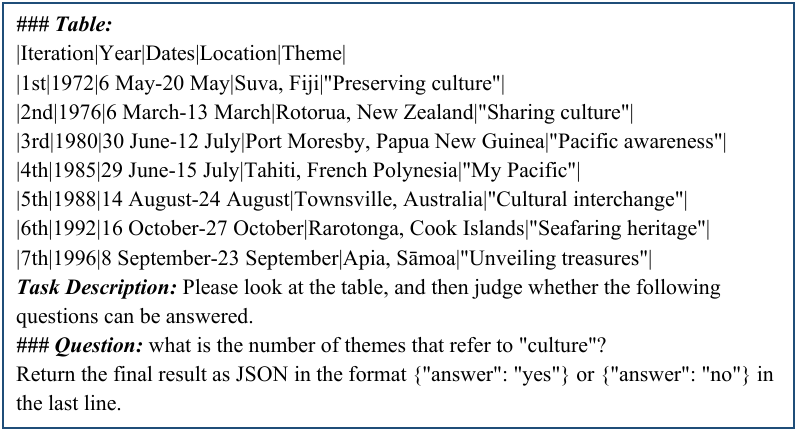}
\vspace{-1mm}
\caption{The prompt for \texttt{Table} subset in ``\textit{negative rejection}'' testbed.} \label{rejection2}
\end{figure}

\begin{figure}[ht]
\centering
\includegraphics[scale=0.56]{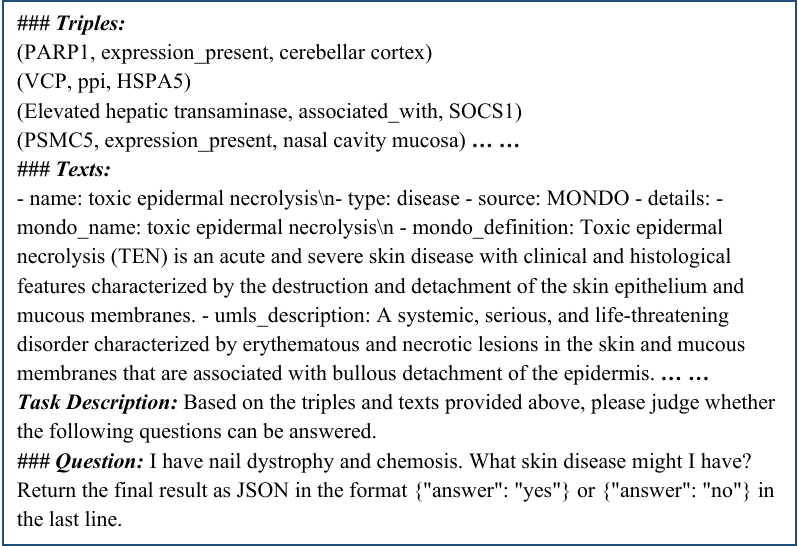}
\vspace{-1mm}
\caption{The prompt for \texttt{KG+Text} subset in ``\textit{negative rejection}'' testbed.} \label{rejection3}
\end{figure}

\begin{figure}[ht]
\centering
\includegraphics[scale=0.56]{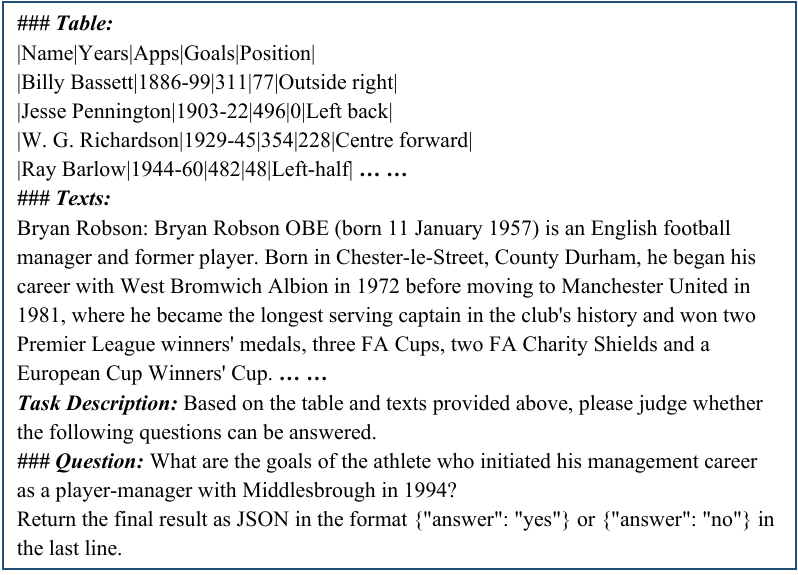}
\vspace{-1mm}
\caption{The prompt for \texttt{KG+Text} subset in ``\textit{negative rejection}'' testbed.} \label{rejection4}
\end{figure}

Fig.~\ref{prompt1},~\ref{prompt2},~\ref{prompt3},~\ref{prompt4} show QA prompt templates for four subsets in \textit{Noise Robustness} testbed, \textit{Order Insensitivity} testbed, and \textit{Information Integration} testbed. Fig.~\ref{rejection1},~\ref{rejection2},~\ref{rejection3},~\ref{rejection4} show prompt templates of \textit{negative rejection} testbed for four subsets.

\end{document}